\documentclass[letterpaper, 10 pt, conference]{ieeeconf}  % Comment this line out if you need a4paper

\IEEEoverridecommandlockouts                              % This command is only needed if 
\usepackage{graphics} % for pdf, bitmapped graphics files
\usepackage{graphicx} % for pdf, bitmapped graphics files

\usepackage{amsmath} % assumes amsmath package installed
\usepackage{amssymb}  % assumes amsmath package installed
\usepackage{multirow}
\usepackage{titlesec}
\titlespacing{\subsection}{0pt}{*0.8}{*0.8}
\titlespacing{\section}{0pt}{*1.0}{*0.8}
\usepackage{cite}
\usepackage{wrapfig}
\usepackage{dcolumn}
\usepackage{makecell}
\usepackage{tabularx}
\usepackage{booktabs}
\usepackage{xcolor}
\usepackage{stfloats}
\newcolumntype{d}[1]{D{.}{.}{#1}}

\title{\LARGE \bf
VISTA: Monocular Segmentation-Based Mapping for Appearance and View-Invariant Global Localization}

\author{Hannah Shafferman$^{1,2,3}$, Annika Thomas$^{1}$, Jouko Kinnari$^{4}$, Michael Ricard$^{3}$, Jose Nino$^{3}$, and Jonathan P.\ How$^{1}$% <-this % stops a space
\thanks{*This work is supported by the Charles Stark Draper Laboratory, Inc. and funded in part by ONR.}% <-this % stops a space
\thanks{$^{1}$Massachusetts Institute of Technology, Cambridge, MA 02139, USA.{\tt\small \{hshaff, annikat, jhow\}@mit.edu}}%  
\thanks{$^{2}$Draper Scholar.}%
\thanks{$^{3}$Charles Stark Draper Laboratory, Inc., Cambridge, MA, USA.}%
\thanks{$^{4}$Saab Finland Oy, 00100 Helsinki, Finland \{jouko.kinnari@saabgroup.com\}}
}

\begin{document}

\maketitle
\thispagestyle{empty}
\pagestyle{empty}

\newcommand{\mytt}[1]{{\color{red}#1}}

% VISTA - View-Invariant Segmentation-based Tracking for Frame Alignment
%%%%%%%%%%%%%%%%%%%%%%%%%%%%%%%%%%%%%%%%%%%%%%%%%%%%%%%%%%%%%%%%%%%%%%%%%%%%%%%%
\begin{abstract}
Global localization is critical for autonomous navigation, particularly in scenarios where an agent must localize within a map generated in a different session or by another agent, as agents often have no prior knowledge about the correlation between reference frames. However, this task remains challenging in unstructured environments due to appearance changes induced by viewpoint variation, seasonal changes, spatial aliasing, and occlusions --- known failure modes for traditional place recognition methods. To address these challenges, we propose VISTA (\underline{V}iew-\underline{I}nvariant \underline{S}egmentation-Based \underline{T}racking for Frame \underline{A}lignment), a novel open-set, monocular global localization framework that combines: 1) a front-end, object-based, segmentation and tracking pipeline, followed by 2) a submap correspondence search, which exploits geometric consistencies between environment maps to align vehicle reference frames. VISTA enables consistent localization across diverse camera viewpoints and seasonal changes, without requiring any domain-specific training or finetuning. We evaluate VISTA on seasonal and oblique-angle aerial datasets, achieving up to a 69\% improvement in recall over baseline methods. Furthermore, we maintain a compact object-based map that is only 0.6\% the size of the most memory-conservative baseline, making our approach capable of real-time implementation on resource-constrained platforms. 

\end{abstract}

%%%%%%%%%%%%%%%%%%%%%%%%%%%%%%%%%%%%%%%%%%%%%%%%%%%%%%%%%%%%%%%%%%%%%%%%%%%%%%%%
\section{Introduction}
Autonomous vehicles operating within GNSS-denied environments, such as urban canyons, forests, indoor facilities, or adversarial settings, require robust localization to navigate in these scenarios \cite{wagner2023robust}. In such settings, vision-based localization is particularly challenging due to significant appearance variability caused by changes in viewpoint, season, lighting, shadows, spatial aliasing, and occlusions. In multi-agent scenarios, global localization --- the task of pose estimation within a previously acquired map without an initial guess \cite{yin2024survey} --- is particularly vital, as it enables agents to align vehicle reference frames for effective information sharing, facilitating coordinated perception and planning. However, agents may observe the environment from  different viewpoints, introducing appearance changes between maps -- a known challenge, even for state-of-the-art methods \cite{thomas2024sosmatch}.

Despite its prevalence in practical scenarios, the off-nadir, or oblique, camera viewpoint --- where the camera is oriented at an angle rather than facing straight downward, depicted in Fig. \ref{fig:pipeline_diagram} --- remains a relatively underexplored challenge. This viewpoint is especially common in heterogeneous teams of UAVs, where agents may have different camera orientations due to domain and task specific sensor configurations. For example, small UAVs often rely on a single forward-facing monocular camera for obstacle avoidance and localization. These oblique viewpoints introduce several unique challenges that hinder mapping, including object distortion, the failure of keypoint-based feature matching methods due to significant appearance changes resulting from different viewing angles, and odometry uncertainty, which can severely impact localization accuracy.

\begin{figure}[t!]
  \centering
    \includegraphics[width=0.48\textwidth]{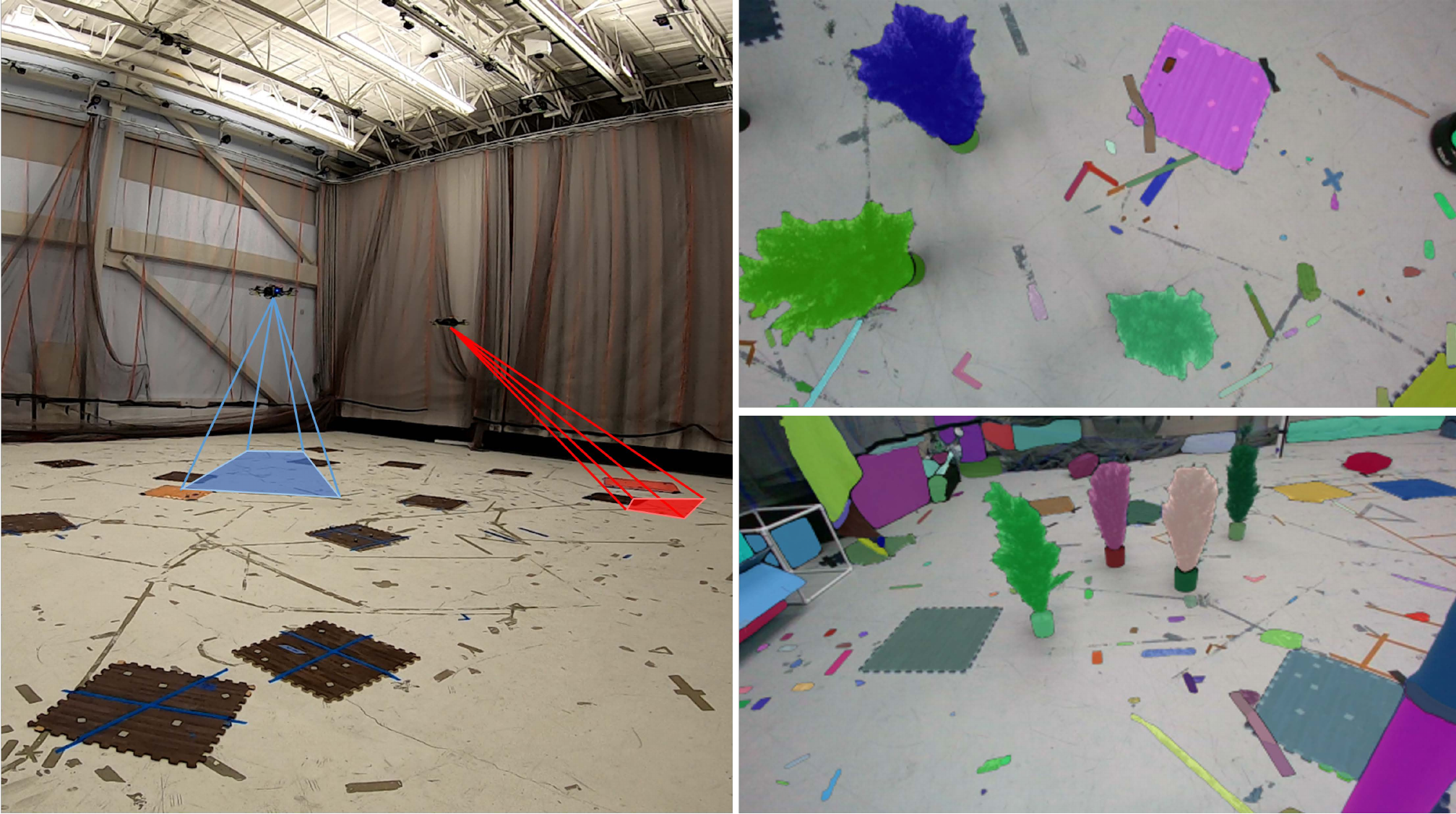}
  \caption{VISTA segments and tracks objects through a video stream from different monocular camera orientations without domain-specific fine tuning in a zero-shot framework to build sparse environment maps. We utilize a data association framework using a geometric submap implementation, which leverages the underlying environment geometry to identify object correspondences between maps.}
  \label{fig:first_page}
\end{figure}

Visual appearance-based approaches, among the most effective methods for localization \cite{galvez2012bags, gawel2018x-view}, rely on detecting and tracking visual features within the operating environment. The appearance of these features can vary significantly across seasons, lighting conditions, and camera viewpoints \cite{lowry2015visual, dusmanu2019d2}. State-of-the-art approaches \cite{lin2020appearance, tian2022kimera, orhan2022semantic, thomas2024sosmatch, peterson2024roman} have addressed aspects of visual appearance changes within the environment, however, they largely focus on nadir or ground-view perspectives. These limitations motivate our focus on developing a framework, which is generalizable to the oblique camera viewpoint without requiring any domain-specific training.

In addition to environmental challenges, the limited computational resources of onboard hardware further constrain the localization solution, as real-time performance is critical for autonomous operation. In large-scale environments with many trackable objects (i.e. a cluttered room, warehouse, or forest environment), maintaining a compact map is essential for scalability. Object-based maps provide this solution, as they offer a lightweight representation compared to dense, feature-based approaches. Even with this compact object-based representation, performing data association over the entire map is computationally expensive and impractical for real-time applications. This challenge motivates our submap approach, which limits the size, and consequently the time, of the correspondence search, while preserving our utilization of the environment geometry. By combining object-based mapping with an efficient submap implementation, we provide a scalable and real-time capable solution for global localization in these visually challenging scenarios.

In this work, we propose VISTA (\underline{V}iew-\underline{I}nvariant \underline{S}egmentation-Based \underline{T}racking for Frame \underline{A}lignment), a monocular, global localization pipeline designed to be robust to extreme environment appearance changes induced by seasonal and camera viewpoint variations. We leverage open-set object-based instance segmentation \cite{kirillov2023segment, ravi2024sam2segmentimages} to generate sparse, viewpoint-invariant, 3D environment representations, depicted in Fig. \ref{fig:first_page}. 
Recent work \cite{thomas2024sosmatch} introduced the mechanism of object segmentation in open-set environments but restricted to the easier nadir viewpoint. VISTA extends that capability by (1) improving the segmentation-based object tracking to support mapping from more challenging oblique viewpoints, which were previously unaddressed, and (2) introducing an efficient, geometrically consistent implementation for submap matching.

In summary, the contributions of this work include: 
\begin{itemize}
    \item A monocular auto-segmentation tracking pipeline that successfully tracks and maps objects from different camera viewpoints. This segment-based object tracker reconstructs a sparse map, containing 3D object locations and their corresponding uncertainties, which is geometrically consistent and can be leveraged for efficient and reliable global localization.
    \item A geometric submap correspondence search method that accounts for object uncertainties and leverages graph theoretic data association, achieving at least a $62\times$ reduction in computation time over most traditional and learning-based baselines, while maintaining significantly smaller maps (0.03\% to 0.6\% the size of the maps generated by the state-of-the-art baselines). Full analysis and explanation are given in Section \ref{experiments_pr}.
    \item Evaluation of the proposed method in the presence of drastic appearance changes within the environment in 1) an unstructured seasonal dataset and 2) challenging oblique viewpoint datasets, achieving a 69\% maximum recall improvement over traditional and learning-based visual place recognition (VPR) baseline methods. 
\end{itemize}

\section{Related Work}
In this section, we review prior work on global localization in unstructured environments and highlight current limitations that motivate our approach for achieving robust performance under extreme environment appearance variation. 

\subsection{Map Representations}
Successful global localization is highly dependent on a descriptive environment representation. Dense map representations (i.e. surfel maps \cite{whelan2015elasticfusion}, 3D Gaussian splats \cite{yugay2023gaussianslam}, NERFs \cite{mildenhall2020nerfrepresentingscenesneural}, or voxel maps \cite{niessner2013real}) encode complex geometric information about the environment but are computationally expensive and are not robust to domain shifts. Sparse maps, on the other hand, are lightweight and scalable environment reconstructions, which represent a scene as a set of 3D points \cite{Cadena_2016}. Sparse maps are commonly either feature-based or object-based representations \cite{yang2019cubeslam, nicholson2018quadricslam}. Feature-based methods store low level distinct object features (i.e. corners, edges). Traditional feature detectors, such as SIFT \cite{lowe2004distinctive}, SURF \cite{bay2008speeded}, and ORB \cite{rublee2011orb}, are widely used to generate unique feature descriptors, but they rely on the assumption during matching that objects appear similarly between frames. Deep learning-based methods \cite{detone2018superpoint, sarlin2020superglue, dusmanu2019d2, sun2021loftr, keetha2023anyloc} have been trained to detect, describe, and match features between frames to improve robustness to appearance changes caused by illumination or motion blur. However, their performance degrades under more extreme appearance changes and when the contents are perceptually similar between scenes \cite{gawel2018x-view, thomas2024sosmatch}. Our work builds upon prior works \cite{thomas2024sosmatch, peterson2024roman}, which leverage object-based environment representations. These object-based maps are lightweight and less sensitive to viewpoint variation and appearance or illumination changes, making them useful for information sharing in multi-agent systems. 

\subsection{Segmentation}
Instance segmentation enables object-based mapping by providing semantics and precise object geometries within unstructured scenes \cite{minaee2021image}. The Segment Anything Model (SAM) \cite{kirillov2023segment} provides zero-shot segmentation for any object in an image. Segment Anything 2 (SAM2) \cite{ravi2024sam2segmentimages} extends this capability to video segmentation by maintaining a memory of segment instances throughout the image sequence, providing robustness even in the presence of occlusions. Our approach builds upon these models by leveraging segment masks as descriptors for robust object-based tracking and re-identification throughout challenging scenes such as in the presence of viewpoint variations, seasonal changes, spatial aliasing, and occlusions.

\subsection{Unstructured Environments}
Many global localization works rely on distinct, repeatable features of urban settings (i.e. roads, street signs, lane markings) to simplify correspondence search for VPR \cite{javanmardi2017towards, kim2021scan, pink2008visual, gawel2018x-view}. In contrast, unstructured environments lack such features and present challenging scenarios due to their irregular nature, repetitive terrain, lighting variation, and changing environmental conditions \cite{ebadi2023present, tian2020search, pritchard2025forestvoenhancingvisualodometry}. These features of unstructured environments make perception difficult and introduce challenges such as perceptual aliasing and occlusions. Open-set operation refers to the capability of a system to handle inputs not previously seen during training. Unstructured environments are typically associated with open-set operation as systems must be robustly designed to handle many unknowns. Our global localization pipeline leverages object-based sparse maps, which generalize well to unstructured environments as opposed to feature-based methods and offer robustness to visual appearance changes common in these scenarios induced by changing environmental conditions.

\section{Method}
\begin{figure*}[ht]
    \centering
    \includegraphics[width=\textwidth, height=55mm]{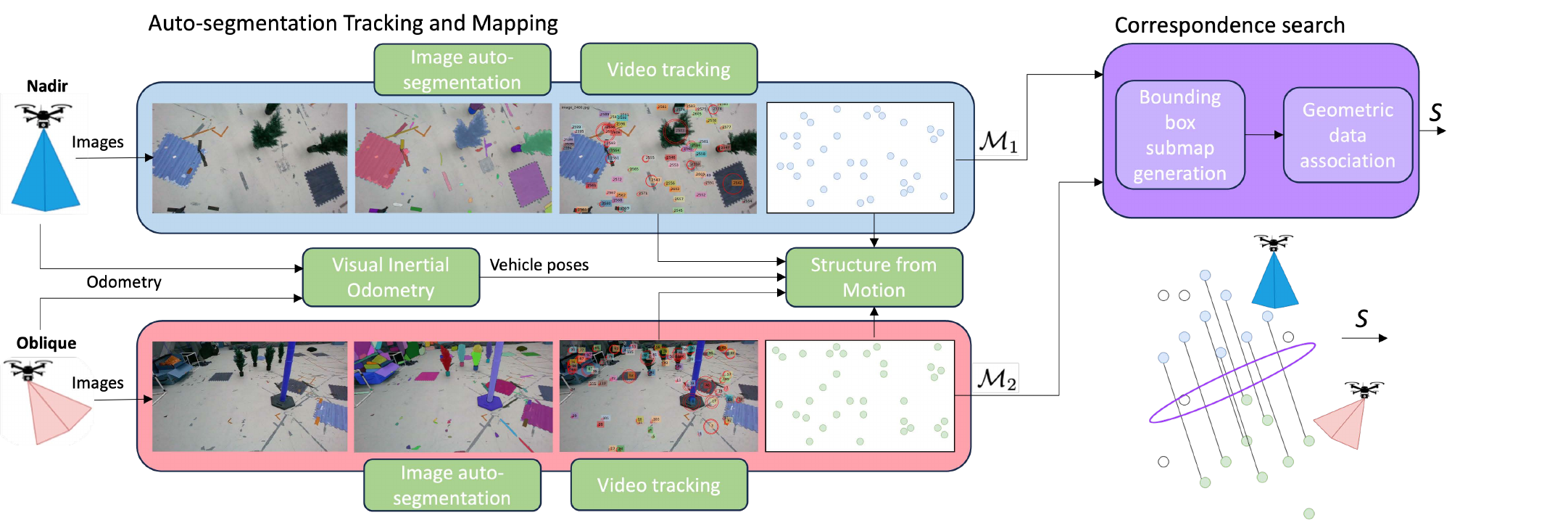}
    \caption{VISTA integrates a novel auto-segmentation object-level tracking pipeline, which tracks objects through camera images and reconstructs a sparse 3D environment representation using the vehicle poses to triangulate object positions and estimate their covariances. Our correspondence search does an all-to-all submap comparison, which leverages geometric consistencies to estimate the frame alignment between vehicles.}
    \label{fig:pipeline_diagram}
\end{figure*}
This section presents an overview of VISTA by first introducing our auto-segmentation object tracking and mapping pipeline followed by a discussion of our submap correspondence search frame alignment step. A visualization of the data flow is depicted in Fig. \ref{fig:pipeline_diagram}.

\subsection{Auto-segmentation Object Tracking}
Our object tracking pipeline relies only on an image sequence as input, in contrast to previous work \cite{thomas2024sosmatch}, which relied on camera pose estimates informed by IMU data to perform fundamental matrix filtering during segment tracking. Our tracker generates a set of object tracks, each consisting of a set of segment masks in every frame for which the object is detected. For vehicle $i$, we define the number of frames in a video sequence as $N_i$ and a group size, $g_i \leq N_i$, as the batch size of frames over which to track and store objects. The video segmentation model \cite{ravi2024sam2segmentimages} requires processing the entire video sequence before object tracking. To support real-time operation, we process images in groups, allowing for batched inference on data from the RGB camera. This group size parameter may be tuned according to memory and computational constraints, enabling flexibility to various hardware platforms. We refer to each group of images by index $l \in \{0, \ldots, \text{ceil}(N_i/g_i)\}$.

For the first image in each group, we utilize a pre-trained image segmentation model \cite{kirillov2023segment} to generate object masks by prompting the model with a $32 \times 32$ grid of points. This grid size provides a dense, uniform sampling over each image, while maintaining computational efficiency. We denote the number of detected objects as $n_{l}$. The model returns a set of binary segment masks, $b_{k,l}$, where $k \in \{1, \ldots, n_{l}\}$, describing the spatial extent of each detected object. We create a chain of model inference by feeding the set of object masks, $b_{k,l}$, to the pre-trained video segmentation model \cite{ravi2024sam2segmentimages}. By leveraging the segment masks directly, we explicitly address the granularity issue \cite{szczepaniak2015interpretation,  li2023semanticsamsegmentrecognizegranularity} that other segmentation pipelines face, ensuring meaningful object-level video segmentation. To account for objects entering and exiting the camera field of view (FOV) as the vehicle travels, we periodically reapply the image segmentation model \cite{kirillov2023segment} when the segmented area of the input image falls below the threshold, $\theta_a = 0.5$. When new objects are detected, they are added to the set of tracked segments. Importantly, the existing tracked objects are not reassigned. Instead, we apply a spatial filtering step, which maintains only new segments that significantly overlap with previously unsegmented regions of the image, ensuring stable tracking throughout the video sequence. Any redundant objects in the map that arise from transitioning between image groups are explicitly resolved in the correspondence search step and do not affect the integrity of the tracking pipeline.

For the $n_l$ objects detected in the image sequence of group $l$, our tracker returns a set of binary segment masks, $\{m_{j}\}$, where $j \in \{0, \ldots, n_l\}$. For each object mask, $m_{j}$, we extract the centroid, $\theta_{j}$, as the 2D detection. We store the set of 2D detections, $\mathcal{D}$, where each entry $\mathcal{D}[t_i]$ is a set of 2D object detections for track index, $t_i \in \{0, \ldots, n_{total}\}$, assuming we have $n_{total}$ objects detected throughout the entire trajectory. We maintain a set, $\mathcal{F}$, where each entry $\mathcal{F}[t_i]$ is a set of frames where each object was detected.

We assume a generic Visual Inertial Odometry (VIO) implementation for providing camera poses $T(t) \in SE(3)$. We utilize a Structure-from-Motion (SfM) triangulation approach to build a 3D environment reconstruction containing estimated object locations and their corresponding uncertainties using only RGB imagery. We construct a factor graph \cite{dellaert2017factor} for each object track $t_i \in \{0, \ldots, n_{total}\}$ with greater than $n_{min} = 3$ detections. For each frame where object $t_i$ was detected, the corresponding camera pose is added as a factor to the graph. For each 2D detection of object $t_i$, we add a projection factor parameterized by the detection and known camera calibration parameters. The factor graph framework \cite{dellaert2017factor} solves for the 3D object position by minimizing the reprojection error. As noted in \cite{dellaert2012factor}, the challenge of SfM is not the computation, rather the data association and the initial guess. Our auto-segmentation tracker robustly handles this challenging data association step and we provide an initial guess using a triangulation. We solve for the 3D positions of each track $t_i$ independently. For objects for which the bundle adjustment fails to converge, we assume it is a dynamic object and discard the object from the map. For each object $t_i$ as seen by vehicle $i$, we generate an environment map, $\mathcal{M}_i$, expressed in the odometry frame of vehicle $i$, containing the estimated 3D positions, $\mu_{t_i}$, and corresponding covariance matrices for all detected objects in the environment. 

\subsection{Submap Correspondence Search}
In multi-vehicle operations within the same environment, establishing correspondences between maps is essential for global localization without an initial pose estimate. The goal of our submap correspondence search is to enable vehicle $i$ to localize within vehicle $j$'s map, $\mathcal{M}_j$, leveraging its own map, $\mathcal{M}_i$, to correctly identify object correspondences.

% While constructing the submaps, we assume no temporal knowledge or trajectory information. Instead, we rely only the 3D object position estimates stored in the map. To partition each environment map into smaller submaps, we implement a sliding window approach. We first compute the Mahalanobis distance of each landmark in the map to the mean of the overall landmark distribution and retain only those within the $\Omega^{th}$ percentile as inliers, adding an extra layer of robustness to poorly estimated object positions. Submaps are defined by a window size, $w$, describing the length and width of each submap, and an overlap parameter, $\alpha$, controlling the step size between the submap centers. The previous work \cite{thomas2024sosmatch} used a naive approach to submap generation, defining subsets of objects by their identification numbers rather than leveraging their geometric locations. In this implementation, depicted in Fig. \ref{fig:submaps}, the submaps represent the true geometry of the overall environment map, where each submap contains the $n_{\text{max}}$ inlier objects whose Euclidean distances are closest to submap center. 
While constructing the submaps, we assume no temporal knowledge or trajectory information. Instead, we rely only the 3D object position estimates stored in the map. To partition each environment map into smaller submaps, we implement a sliding window approach. We first compute the Mahalanobis distance of each landmark in the map to the mean of the overall landmark distribution and retain only those within the $\Omega^{th}$ percentile as inliers, adding an extra layer of robustness to poorly estimated object positions. Submaps are defined by a window size, $w$, describing the length and width of each submap, and an overlap parameter, $\alpha$, controlling the step size between the submap centers. The previous work \cite{thomas2024sosmatch} used a naive approach to submap generation, defining subsets of objects by their identification numbers rather than leveraging their geometric locations. In this implementation, the submaps represent the true geometry of the overall environment map, where each submap contains the $n_{\text{max}}$ inlier objects whose Euclidean distances are closest to submap center. 
% \begin{figure}[thpb]
% \begin{wrapfigure}{R}{0.25\textwidth}
%     \centering
%     \includegraphics[width=0.25\textwidth]{figures/submap_visualization.pdf}
%     \caption{Visualization of the submaps for an all-to-all comparison search between maps, $\mathcal{M}_i$ and $\mathcal{M}_j$.}
%     \label{fig:submaps}
% \end{wrapfigure}

\begin{figure}[th]
    \centering
    \includegraphics[width=0.5\textwidth, height=18mm]{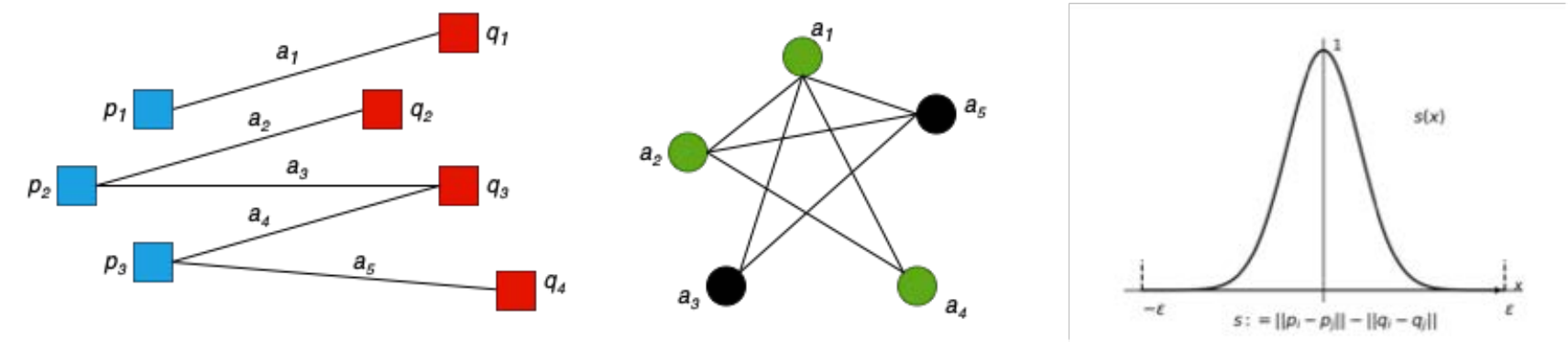}
    \caption{Consistency graph formulation for example 3D submap registration problem. Candidate associations $a_1, \ldots, a_5$ between two submaps (left) become vertices of consistency graph, $\mathcal{G}$, with edges representing their geometric consistency (middle). The geometrically consistent set is highlighted in green. Edges of $\mathcal{G}$ are weighted based on a pairwise consistency scoring function $s(x)$ (right).}
    \label{fig:clipper}
\end{figure}

We utilize a robust data association framework \cite{lusk2021clipper, lusk2024clipper}, to compute 3D point associations between submaps. We restructure the problem into a graph formulation to solve for the densest geometrically consistent clique within the subgraph, depicted in Fig. \ref{fig:clipper}. Geometric consistency is determined using pairwise distances between points in each point cloud since rotation and translation transformations preserve distance. Without the presence of noise, two points in $\mathcal{M}_1$, say $p_i$ and $q_i$, being mapped to points in $\mathcal{M}_2$, $p_i'$ and $q_i'$ respectively, are geometrically consistent if $\|p_i - q_i\| = \|p_i' - q_i'\|$. We define $x := \|p_i - q_i\| - \|p_i' - q_i'\|$ as the pairwise distance difference between pairs of points. However, in the presence of noise, a scalar function, $s(x)$, is used to weight edges in $\mathcal{G}$ by, 
\begin{equation}
    s(x) = \begin{cases}
        \text{exp}\big(-\frac{1}{2}\frac{x^2}{\sigma^2} \big) & \text{if } |x| \leq \epsilon \\
        0 & \text{if } |x| > \epsilon
    \end{cases}
\end{equation}
The function $s(x)$ is bounded by a tunable parameter, $\epsilon$, and parameterized by an expected noise parameter, $\sigma$. This allows for some noise tolerance in the point cloud correspondence search while weighing the associations by the expected noise.

These weights are stored in a weighted affinity matrix $A$, where $A_{p,p'} = s(x)$ and $A_{p,p} = 1$. The densest geometrically consistent clique then follows by solving the optimization, 
\begin{equation}
    \begin{aligned}
        \max_{u \in \{0,1\}^n} \quad & \frac{u^T A u}{u^T u} \\
        \text{subject to} \quad & u_p u_{p'} = 0, \quad \text{if } A_{p,p'} = 0, \quad \forall p, p',
    \end{aligned}
\end{equation}
where $u_p$ is 1 when an association is identified as an inlier and 0 when an association is deemed an outlier.

\begin{wrapfigure}{R}{0.4\columnwidth}
% \begin{figure}[thpb]
    \centering
    \includegraphics[width=0.4\columnwidth]{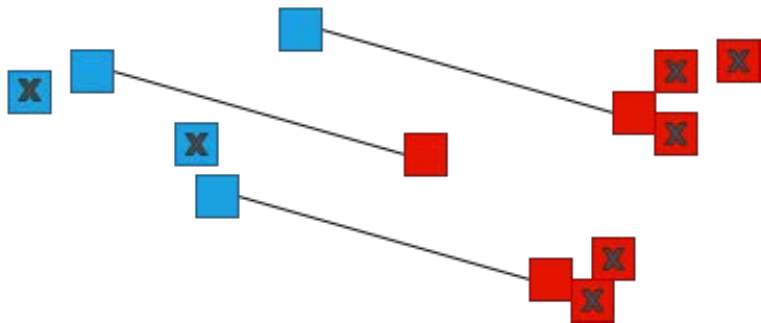}
    \caption{Visualization of minimum distance parameter, $\gamma$. Correspondences between the blue and red point clouds are drawn. Correspondences disqualified by $\gamma$ are marked.}
    \label{fig:mindist}
\end{wrapfigure}

A parameter, $\gamma$, depicted in Fig. \ref{fig:mindist}, implicitly addresses segmentation inconsistencies introduced during data association. Specifically, $\gamma$ enforces a minimum distance between any two matched points within the same map, thus handling any duplicate objects introduced during tracking. This approach is preferred over clustering techniques, which can struggle to distinguish between duplicate objects and noise. 

Each submap correspondence search returns the largest geometrically consistent set of object pairs, denoted as $\mathcal{S}$, between two submaps. Using Arun's method \cite{arun1987least}, we estimate the relative rotation and translation between the object pairs, which we consider a candidate global transformation between vehicle $i$ and vehicle $j$'s reference frames. We prune any candidate transformations that have a pitch or roll greater than $\theta_{RP}$ degrees due to dynamic infeasibility. We use the cardinality of $\mathcal{S}$ greater than a threshold $S_{\text{max}}$, $|\mathcal{S}|>S_{\text{max}}$, as a success metric in the following experiments. We vary the $S_{\text{max}}$ threshold, to balance precision and recall for our approach.

\section{Experiments}
We evaluate our performance on both the B\aa tvik seasonal dataset \cite{thomas2024sosmatch} and a Highbay dataset, comprised of two challenging video sequences captured from both the nadir and oblique viewpoints collected in a large indoor flight testing facility. We compare the precision, recall, map size, and search time of VISTA to five baseline approaches.

\subsection{Baselines}
ORB \cite{rublee2011orb} is a commonly used visual feature detection method, typically combined with RANSAC \cite{fischler1981random} for matching feature descriptors and rejecting outlier detections. We implement an ORB+RANSAC baseline method, which detects a maximum of 500 ORB features in each keyframe. We utilize a RANSAC back-end to identify k-nearest neighbor correspondences $(k = 2)$ between ORB features and retain the feature matches that survive Lowe's ratio test \cite{lowe2004distinctive}. 

We compare VISTA to two state-of-the-art learning-based feature matching approaches. We benchmark our results against LoFTR \cite{sun2021loftr} as well as SuperPoint \cite{detone2018superpoint} feature detection front-end combined with SuperGlue \cite{sarlin2020superglue} feature matching back-end. For both methods we use pretrained outdoor weights for the B\aa tvik seasonal dataset \cite{thomas2024sosmatch} and pretrained indoor weights for the Highbay datasets. 

We also compare against AnyLoc \cite{keetha2023anyloc}, a VPR approach which generates global image descriptors rather than local feature descriptors. The authors claim that AnyLoc universally performs well in challenging VPR conditions (i.e. different environments, lighting conditions, and perspectives), making this a strong benchmark comparison for our method. For our comparison, we follow the AnyLoc parameters that achieve best performance. We define a DINOv2 extractor at layer 31 with facet \textit{value} and load the VLAD vocabulary with 32 cluster centers. We generate global descriptors for each keyframe and compute the cosine similarity of the global descriptors for each image pair in the datasets.

The previously published work, SOS-Match \cite{thomas2024sosmatch} addressed localization under appearance changes induced by seasonal variation. SOS-Match outperformed state-of-the-art traditional and learning-based baseline approaches utilizing nadir imagery, but has difficulties generalizing to imagery taken from the oblique camera angle. 

\subsection{Performance metrics}
We evaluate VISTA by computing precision and recall. For each pair of submaps being considered by the correspondence search, we compute the IoU of the 3D point clouds, and assume that for each pair of submaps with enough overlap, $\text{IoU} > \theta_o$, we expect a correct transformation to be returned. A correct transformation is defined to have roll and pitch less than $\theta_{RP}$ degrees, due to the dynamic feasibility of the transformation, a yaw less than $\theta_Y$, and a translation less than $T_{\text{max}}$ meters, since we are mapping the environments in the same reference frame for comparison. We define \textit{hypothesized matches} to mean candidate transformations that, from the context of the algorithm, seem correct based on having a small roll and pitch, less than $\theta_{RP}$ degrees. In the context of our scenario, {\em precision} is defined as the fraction of true correct transformations out of the total number of hypothesized transformations. {\em Recall} is defined as the fraction of true correct transformations with $\text{IoU} > \theta_o$, such that we expect a correspondence out of the total number of submap pairs with sufficient overlap.

We vary the cardinality threshold, $S_{\text{max}}$, defining the minimum number of detected point correspondences between submaps, to produce precision and recall results for our approach and SOS-Match \cite{thomas2024sosmatch}. For the ORB+RANSAC baseline, we similarly vary the number of feature correspondences detected between keyframes. For SuperPoint+SuperGlue and LoFTR, we vary the minimum match confidence threshold for image pairs. For AnyLoc, we vary the minimum cosine similarity score. 

We record the average runtime of one correspondence search step for VISTA compared to each baseline method. This correspondence search time is reflective of the time required to determine if a localization event has occurred in real-time settings. These results are computed using a AMD Ryzen Threadripper 3960X CPU with 128 GB RAM. The results for SuperPoint+SuperGlue, LoFTR and AnyLoc, which require a GPU for correspondence search, are computed using a NVIDIA RTX 3090 GPU. 

\subsection{Datasets}
\begin{figure}[t!]
\vspace{10pt}
\includegraphics[width=0.45\textwidth]{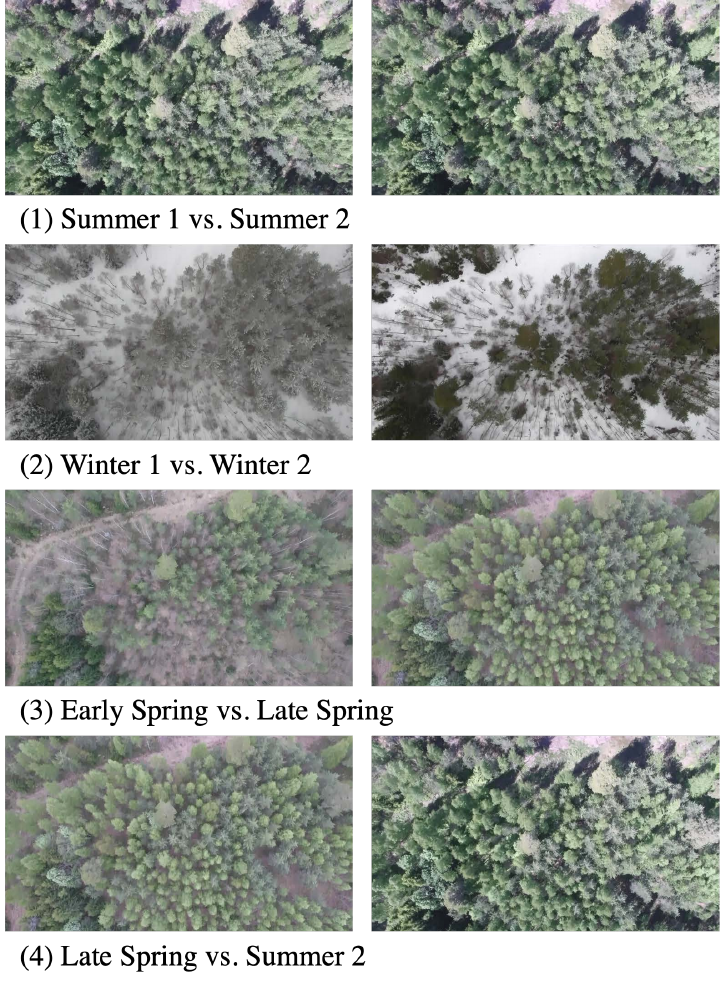}
\centering
\vspace{-10pt}
\caption{Example imagery from the B\aa tvik seasonal dataset \cite{thomas2024sosmatch} demonstrating each of our four seasonal experiments. Cross-seasonal imagery highlights the appearance changes induced by changes in foliage, sharpness of shadows, and snow coverage.}
    \label{fig:batvik_sample_images}
\end{figure}

We leverage the B\aa tvik seasonal dataset \cite{thomas2024sosmatch}, consisting of six 3.5 km UAV flights over a dense forest region recorded over the course of a year. This dataset presents a challenging scenario for long-term VPR due to the drastic appearance changes between drone flights induced by seasonal variation. Fig. \ref{fig:batvik_sample_images} depicts sample nadir imagery from our four cross-seasonal comparisons. These images highlight the significant appearance changes in the environment due sharpening of shadows, changes in lighting, foliage, and snow coverage between seasons. See \cite{thomas2024sosmatch} for further detail about this dataset. 

We also test VISTA on a Highbay dataset to evaluate its robustness to variations in camera perspective. This dataset consists of two experiments, \textit{Experiment 0} and \textit{Experiment 1}, each containing imagery recorded from both the nadir and oblique viewpoints. The \textit{Experiment 0} environment consists of 18 flat pads randomly placed on the ground, with 2841 and 2684 total tracked objects from the nadir and oblique viewpoints, respectively. \textit{Experiment 1} builds upon this setup, adding small trees, tall poles, boxes, and backpacks randomly placed, resulting in a total of 36 large objects. In this experiment, 576 and 706 objects are tracked from the nadir and oblique viewpoints, respectively. Notably, we follow different trajectories between the nadir and oblique viewpoint sequences. As a result, we evaluate not only VISTA's mapping capability under different camera configurations but also its robustness to objects observed from significantly different perspectives -- a known failure mode for many VPR methods, as objects can appear substantially altered depending on viewpoint. We summarize the hyperparameters in Table \ref{hyperparameters_table}.

\subsection{Seasonal Variation} 
We first evaluate VISTA in a scenario where two agents, both equipped with nadir-facing cameras, traverse the same trajectory under varying seasonal conditions. We assess performance across four experiments shown in Fig. \ref{fig:batvik_sample_images}: Summer 1 vs. Summer 2, Winter 1 vs. Winter 2, Early Spring vs. Late Spring, and Late Spring vs. Summer 2. Throughout these experiments, visual discrepancies progressively increase, posing more challenging scenarios for VPR methods.

\begin{table}[t!]
\vspace{10pt}
\caption{Hyperparameters}
\vspace{-15pt}
\label{hyperparameters_table}
\begin{center}
\resizebox{\columnwidth}{!}{
\begin{tabular}{p{.55cm} p{1.25cm} p{2.25cm} p{5.7cm}}
\hline
\textbf{Exp.} & \textbf{Parameter} & \textbf{Value} & \textbf{Description} \\
\hline \hline
0 / 1 & \centering $g_i$ & 50 / 50 & Image group batch size\\
0 / 1 & \centering $\Omega$ & [95/95] / [85/80] & Percentile inlier detections to keep [nadir/oblique]\\
% \hline
0 / 1  & \centering $w$ & 2.0 / 2.0 & Submap window size\\
% \hline
0 / 1 & \centering $\alpha$ & 1.0 / 1.0 & Submap overlap parameter\\
% \hline
0 / 1 & \centering $n_{\text{max}}$ & 50 / 50 & Max. $\#$ of objects in submap\\
% \hline
0 / 1 & \centering $\sigma$ & 0.05 m / 0.05 m & Pairwise consistency expected noise\\
% \hline
0 / 1 & \centering $\epsilon$ & 0.1 m /  0.1 m & Weighting function $s(x)$ cutoff\\
% \hline
0 / 1 & \centering $\gamma$ & 0.1 m / 0.2 m & Min. distance between correspondences\\
% \hline
0 / 1  & \centering $S_{\text{max}}$ & 4 / 4 & Initial min. $\#$ correspondences\\
% \hline
0 / 1  & \centering $\theta_o$ & 0.667 / 0.667 & Submap overlap threshold\\
% \hline
0 / 1 & \centering $\theta_{RP}$ & 10 deg / 6 deg & Roll/pitch threshold\\
% \hline
0 / 1 & \centering $\theta_{Y}$ & 30 deg / 30 deg & Yaw threshold\\
% \hline
0 / 1 & \centering $T_{\text{max}}$ & 1.5 m / 1.5 m & Max. translation threshold\\
\hline

\end{tabular}
}
\end{center}
\end{table}

\begin{figure}[t!]
    \centering
    \includegraphics[width=0.5\textwidth]{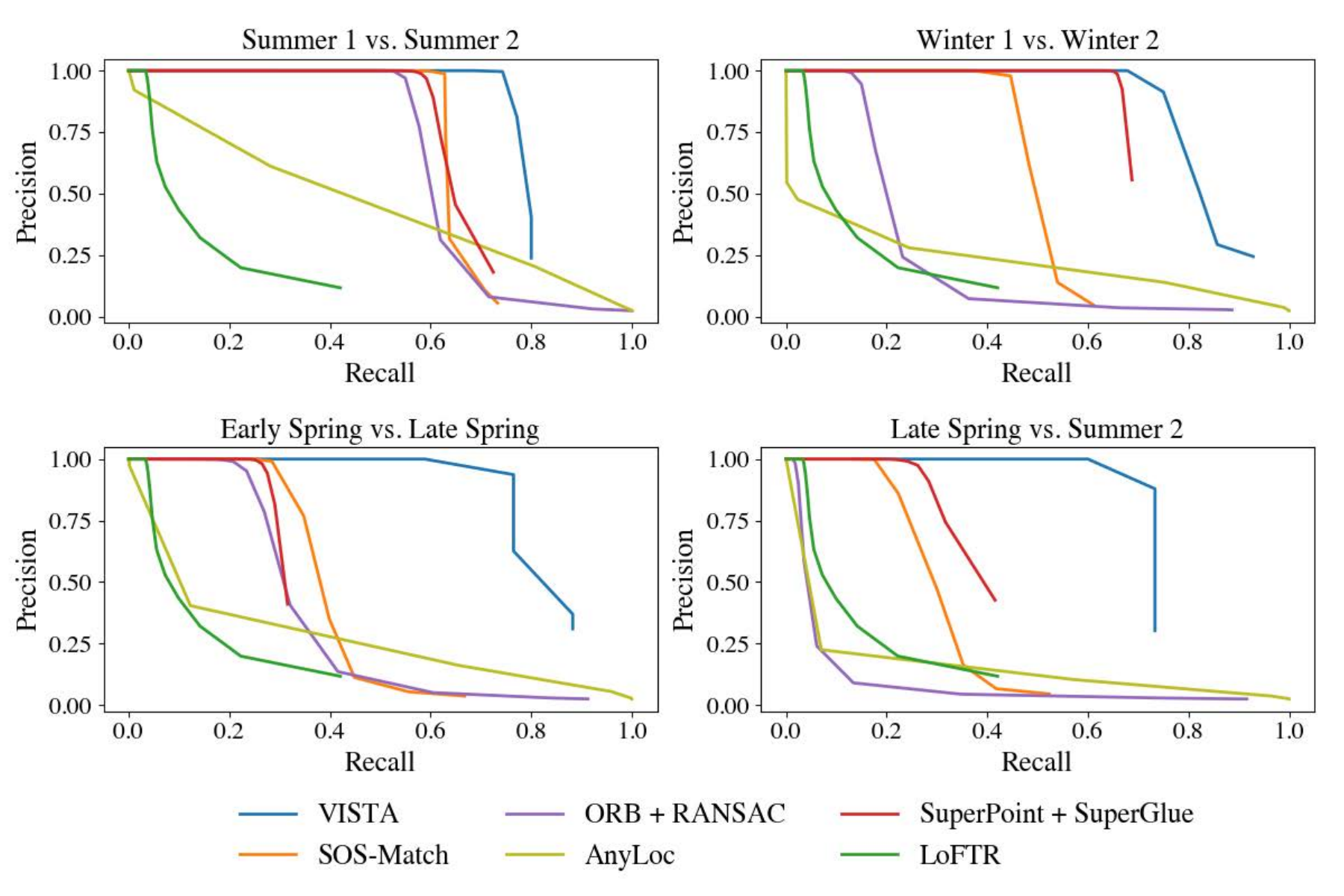}
\vspace{-20pt}
\caption{Precision vs. recall curves with increasing visual discrepancy between flights induced by seasonal variation.}
    \label{fig:sos-match-comparison}
\end{figure}

We compare our results to the baseline methods using a camera projection to estimate the visible region at each keyframe, as described in \cite{thomas2024sosmatch}. We compute the IoU of the submaps being compared to determine if there if sufficient overlap for a valid correspondence. The precision and recall curves demonstrating the performance of VISTA compared to the baseline approaches is shown in Fig. \ref{fig:sos-match-comparison}. 

VISTA outperforms all baseline methods in each of the four experiments. In the most challenging scenarios, with greater visual discrepancy between seasons, we can observe that the performance of all baseline methods significantly degrades. However, VISTA maintains consistent performance throughout the four scenarios and, as a result, shows a large performance improvement in these challenging cases. It is important to note that all of the baseline methods discard sections of the trajectory that fly over water due to the lack of visual features over these regions. The reported results for VISTA do not discard these sections of the trajectory, which further highlights the improved performance and robustness to appearance changes within the environment.

In Table \ref{batvik_recall_at_table}, we report the recall values at $100\%$, $90\%$, $80\%$ precision for these seasonal variation scenarios. As precision decreases, recall will increase, however, it is impractical to utilize a system with very low precision, so we investigate the recall performance only when precision is high. As shown in Table \ref{batvik_recall_at_table}, VISTA achieves the highest recall values compared to the baseline approaches for all of the experiments and all precision thresholds with a maximum 42.8\% improvement over the second best result.

\begin{table*}[bh!]
\caption{Recall @ Precision, Nadir vs. Nadir, B\aa tvik Seasonal Dataset, \textbf{Best} result for each precision threshold highlighted, \underline{Second best} result underlined}
\vspace{-10pt}
\label{batvik_recall_at_table}
\begin{center}
\resizebox{2\columnwidth}{!}{
\begin{tabular}{|c||ccc|ccc|ccc|ccc|}
\hline
 & \multicolumn{3}{c|}{\textbf{Summer 1 vs. Summer 2}} & \multicolumn{3}{c|}{\textbf{Winter 1 vs. Winter 2}} & \multicolumn{3}{c|}{\textbf{Early Spring vs. Late Spring}} & \multicolumn{3}{c|}{\textbf{Late Spring vs. Summer 2}} \\
% \hline \hline
Methods & R@100 & R@90 & R@80 & R@100 & R@90 & R@80 & R@100 & R@90 & R@80 & R@100 & R@90 & R@80 \\
% \hline \hline
\hline
SOS-Match \cite{thomas2024sosmatch} & 25.6 & \underline{62.9} & \underline{62.9} & 12.4 & 45.4 & 46.5 & 17.5 & \underline{31.0} & \underline{33.9} & 12.7 & 20.9 & 23.5 \\
% \hline
ORB \cite{rublee2011orb} + RANSAC \cite{fischler1981random} & 46.9 & 55.9 & 57.3 & 9.9 & 15.5 & 16.5 & 11.1 & 24.5 & 26.6 & 1.2 & 2.4 & 2.8 \\
% \hline
AnyLoc \cite{keetha2023anyloc} & 0.0 & 2.9 & 11.7 &  0.0 & 0.03 & 0.06 & 0.0 & 1.7 & 3.8 & 0.0 & 0.9 & 1.8 \\
% \hline
SuperPoint \cite{detone2018superpoint} + SuperGlue \cite{sarlin2020superglue} & \underline{52.8} & 60.3 & 61.3 & \underline{62.6} & \underline{66.9} & \underline{67.5} & \underline{22.4} & 28.0 & 29.1 & \underline{17.8} & \underline{28.5} & \underline{30.5} \\
% \hline
LoFTR \cite{sun2021loftr} & 3.2 & 4.1 & 4.5 & 3.2 & 4.1 & 4.5 & 3.2 & 4.1 & 4.5 & 3.2 & 4.1 & 4.5 \\
\textbf{VISTA} & \textbf{68.6} & \textbf{75.8} & \textbf{77.2} & \textbf{67.9} & \textbf{75.2} & \textbf{77.0}  & \textbf{58.8} & \textbf{76.5} & \textbf{76.5}  & \textbf{60.0} & \textbf{71.0} & \textbf{73.3}  \\
\hline
\end{tabular}
}
\end{center}
\end{table*}

\subsection{Oblique Camera Viewpoint} \label{experiments_pr}
We also evaluate VISTA on the Highbay datasets in a scenario where two agents, equipped with nadir and oblique camera configurations respectively, operate within the same environment and aim to localize within each other's maps.

In Table \ref{highbay_nn_no_comparison}, we record the recall performance at $100\%$, $90\%$, $80\%$ precision thresholds, first comparing between maps generated using only the nadir-facing imagery, followed by a comparison using the nadir-oblique configuration. We observe that in the nadir-only scenario, VISTA achieves high recall, 83.5\%\,-\,90\%, for all precision thresholds. In the challenging, nadir-oblique camera configuration case, the recall performance degrades from the nadir-only scenario. We compare the performance of VISTA in this challenging case to the five baseline approaches seen in Table \ref{tab:highbay_recall_at}. Similarly to Table \ref{batvik_recall_at_table}, we record the recall values of VISTA compared to five baseline methods at $100\%$, $90\%$, $80\%$ precision. A comparison of map size and runtime is shown in Table \ref{map_size_runtime_table}. VISTA outperforms all baseline approaches at each precision threshold in both datasets. We observe a minimum 9\% and maximum 33\% improvement over the second best results. Our method does so while maintaining a map size that ranges from 0.03\% to 0.6\% the size of baseline methods. Additionally, VISTA outperforms all baseline methods (except AnyLoc \cite{keetha2023anyloc}) in terms of runtime. However, while AnyLoc runs faster than VISTA, it does so with a map size \textbf{over 1200$\times$ larger}, making communication of these maps between vehicles impractical due to the limited bandwidth constraints imposed by typical robot hardware.

Further, SOS-Match \cite{thomas2024sosmatch}, struggles to maintain consistent feature tracks from the oblique camera viewpoint due to increased depth uncertainty and changes in object scale, and therefore segmentation-mask size, used its data association technique. In this oblique angle scenario, the inconsistent object tracking leads to poor 3D object position estimates, degrading mapping performance. This, in turn, prevents accurate reconstruction of the environment's geometry and hinders the identification of submap correspondences necessary for global localization.

\subsection{Large Scale Simulation Environment}
We test VISTA using the Highbay datasets due to a lack of publicly available large-scale oblique viewpoint aerial datasets, but we acknowledge there is a large difference in scale between the B\aa tvik seasonal dataset \cite{thomas2024sosmatch} and the Highbay datasets. Therefore, we test VISTA in a large-scale simulated environment \cite{shah2018airsim} to verify consistent performance regardless of scale. In Table \ref{tab:airsim_table}, we report recall at $100\%$, $90\%$, $80\%$ precision thresholds, for the nadir-only and nadir-oblique scenarios -- the same analysis as for the Highbay datasets. The results in Table \ref{tab:airsim_table} achieve comparable performance to the Highbay results reported in Table \ref{highbay_nn_no_comparison}, with near perfect recall in the easy nadir-only case and slightly degraded recall in the challenging oblique viewpoint scenario. This analysis indicates that VISTA achieves global localization independent of environment scale and therefore it is reasonable to leverage the Highbay datasets for our oblique viewpoint analysis.

\begin{table}[t!]
\vspace{10pt}
\caption{Recall @ Precision, Highbay Dataset Performance}
\vspace{-15pt}
\label{highbay_nn_no_comparison}
\begin{center}
\resizebox{\columnwidth}{!}{
\begin{tabular}{|c||ccc|ccc|}
\hline
 & \multicolumn{3}{c|}{\textbf{Experiment 1}} & \multicolumn{3}{c|}{\textbf{Experiment 0}}\\
% \hline \hline
 & R@100 & R@90 & R@80 & R@100 & R@90 & R@80 \\
\hline 
Nadir/Nadir & 83.5 & 88.6 & 88.6 & 85.0 & 90.0 & 90.0 \\
% \hline
Nadir/Oblique & 33.3 & 33.3 & 40.1 & 16.7 & 36.0 & 40.5 \\
% \hline
% Oblique/Oblique & - / - / - & - / - / - \\
\hline
\end{tabular}
}
\end{center}
\end{table}

\begin{table}[t]
\caption{Recall @ Precision, Highbay Nadir vs. Oblique, \textbf{Best} and \underline{Second} best performance marked}
\vspace{-5pt}
\centering
\small
\setlength{\tabcolsep}{3pt}
\label{tab:highbay_recall_at}
% \resizebox{\columnwidth}{!}{
% \begin{tabularx}{\textwidth}{l *{2}{c}}
\resizebox{\columnwidth}{!}{
    \begin{tabular}{lcccccc}
\toprule
& \multicolumn{3}{c}{\textbf{Experiment 1}} & \multicolumn{3}{c}{\textbf{Experiment 0}}  \\
% \hline \hline
Method & R@100 & R@90 & R@80 & R@100 & R@90 & R@80 \\
\midrule
SOS-Match \cite{thomas2024sosmatch} & - & 0.4 & 0.8 & \underline{8.0} & 8.0 & 8.0 \\
ORB \cite{rublee2011orb} + RANSAC \cite{fischler1981random} & - & 0.2 & 4.0 & 0.0 & 4.0 & 10.0 \\
AnyLoc \cite{keetha2023anyloc} & 0.03 & 1.0 & 5.0 & 0.0 & 0.3 & 0.5  \\
SuperPoint \cite{detone2018superpoint} + SuperGlue \cite{sarlin2020superglue} & 0.1 & \underline{11.0} & \underline{18.0} & 1.0 & \underline{18.0} & \underline{25.0} \\
LoFTR \cite{sun2021loftr} & \underline{0.3} & 3.0 & 4.0 & 0.03 & 6.0 & 7.0  \\
\textbf{VISTA} & \textbf{33.3} & \textbf{33.3} & \textbf{40.1} & \textbf{17.0} & \textbf{36.0} & \textbf{41.0} \\
\bottomrule
\end{tabular}
}
\end{table}

\subsection{Ablation}
Our auto-segmentation tracking pipeline leverages a segmentation model \cite{kirillov2023segment} to identify binary segment masks for all objects in an image frame. By leveraging these segment masks directly, our approach creates a chain of model inferences to track all objects across frames \cite{ravi2024sam2segmentimages}. The video segmentation model \cite{ravi2024sam2segmentimages} accepts points, bounding boxes, or masks (what we propose herein) as input to define the object of interest for tracking. In Table \ref{highbay_ablation_table} we evaluate the impact of different input methods on object tracking by altering our pipeline to specify the object of interest using each of the three methods. We evaluate the impact on localization by recording recall at $100\%$, $90\%$, $50\%$ precision thresholds. We lower the minimum precision threshold to accommodate the degraded performance of different prompting methods, which do not reach higher precision values. 

The results in Table \ref{highbay_ablation_table} demonstrate that directly using the descriptor information contained in the segment masks connects the most relevant information for tracking as opposed to object bounding boxes or center points. The bounding box method fails to achieve either high recall or high precision. The point method can achieve high precision, but that is at the cost of poor recall. 

At 100\% precision on the Experiment 0 dataset, the point method achieves slightly higher recall than our segment mask approach, though it is quickly surpassed by our method at 90\% precision. We attribute this to the Experiment 0 environment, which consists of only flat pads and lacks object diversity. Our method, which leverages the segment masks directly, captures the true object geometry, enabling superior performance in the more complex environments, as shown by the localization results on the Experiment 1 dataset.

% \newcolumntype{d}[1]{D{.}{.}{#1}}
\begin{table}[t!]
\vspace{10pt}
\caption{Map size and match search time comparison. \textbf{Best} and \underline{second} best performance are marked}
\vspace{-15pt}
\label{map_size_runtime_table}
\begin{center}
\resizebox{\columnwidth}{!}{
\begin{tabular}{|l||c|c|c|}
\hline
Implementation & Map size (Mb) & \multicolumn{2}{c|}{Comparison} \\
&&runtime (s) & std\\ 
\hline 
SOS-Match \cite{thomas2024sosmatch} &  \underline{0.62} & 5.24 & 1.37 \\
ORB \cite{rublee2011orb} + RANSAC \cite{fischler1981random} &  101.4 & 4663.3 & 15.3\\
AnyLoc \cite{keetha2023anyloc} & 708.6 & \textbf{0.12} & \textbf{0.003}\\
SuperPoint \cite{detone2018superpoint} + SuperGlue \cite{sarlin2020superglue} & 1811.0 & 67.22 & 0.72\\
LoFTR \cite{sun2021loftr} & 1102.3 & 245.58 & 0.08 \\
\textbf{VISTA} & \textbf{0.59} & \underline{1.08} & \underline{0.67}\\
\hline
\end{tabular}
}
\end{center} 
\end{table}

\begin{table}[th]
\caption{Recall @ Precision, Large-scale Simulated Environment}
\vspace{-5pt}
\centering
\scriptsize
\setlength{\tabcolsep}{3pt}
\label{tab:airsim_table}
% \resizebox{\columnwidth}{!}{
% \begin{tabularx}{\textwidth}{l *{2}{c}}
% \resizebox{\columnwidth}{!}{
    \begin{tabular}{lccc}
\toprule
& \multicolumn{3}{c}{\textbf{AirSim \cite{shah2018airsim}}}\\
% \hline \hline
Method & R@100 & R@90 & R@80 \\
\midrule
Nadir/Nadir & 73.1 & 98.5 & 98.5 \\
Nadir/Oblique & 1.1 & 55.8 & 55.8 \\
% Nadir/Nadir & 98.2 & 99.8 & 99.8 \\
% Nadir/Oblique & 30.0 & 100.0 & 100.0 \\
\bottomrule
\end{tabular}
% }
%\end{table}
%
%\begin{table}[th]
\caption{Auto-Segmentation Tracker Ablation}
\vspace{-15pt}
\label{highbay_ablation_table}
\begin{center}
\resizebox{\columnwidth}{!}{
\begin{tabular}{|c||ccc|ccc|}
\hline
 & \multicolumn{3}{c|}{\textbf{Experiment 1}} & \multicolumn{3}{c|}{\textbf{Experiment 0}} \\
% \hline \hline
Prompting Method & R@100 & R@90 & R@50 & R@100 & R@90 & R@50 \\
\hline 
% \hline
Point & 0.0 & 4.0 & 20.0 & \textbf{33.3} & 33.3 & 33.3 \\
% \hline
Bounding Box &  - & - & 5.3 & 0.0 & 0.0 & 39.6 \\
\textbf{Mask (ours)} & \textbf{33.3} & \textbf{33.3} & \textbf{63.2} & 16.7 & \textbf{36.0} & \textbf{50.0} \\

\hline
\end{tabular}
}
\end{center}
\end{table}

\section{Discussion}
This work highlights the power of combining deep learning models with sparse geometric front-end mapping methods to achieve our goal of localization across a team of vehicles operating in environments with significant appearance changes. Our auto-segmentation tracker identifies object segments using only image sequences as input to reconstruct 3D object-based maps. This offers two key advantages as opposed to previous work \cite{thomas2024sosmatch}. First, our approach eliminates the need for size descriptors, which we found to degrade quality, particularly in oblique view scenarios. Second, our approach removes the dependence on camera pose estimates in the tracking front-end, enhancing robustness by decoupling data association from potential pose estimation errors.

Our 3D object-based maps accurately preserve the environment geometry, enabling object correspondences to be identified between vehicle maps. We highlight that VISTA is not only capable of geometrically consistent mapping under extreme visual appearance changes induced by seasonal variation or different camera orientations, but also robust to changes in viewpoint, which we demonstrate in our Highbay experiments where vehicles reconstruct the environment following different trajectories. We ensure that our environment map is lightweight and runtime is low enough that VISTA can be used for localization or loop closure detection for teams of autonomous vehicles operating in real-time. 

Our method does not attempt to identify when the vehicle has returned to a previously mapped location and does not explicitly handle duplicate objects within the map. Rather, we implicitly address these inconsistencies during the submap correspondence search when computing the largest pairwise, geometrically consistent set of objects, we set a $\gamma$ threshold, which prevents multiple points mapped within the threshold distance from being included in the maximum pairwise consistent set. In future work, we plan to utilize our method for loop closure detection within a SLAM framework. In addition, we plan to incorporate semantic information into our long-term localization framework to further disambiguate between submaps in these challenging monocular VPR scenarios under drastic environment appearance changes. 
 
\section{Conclusion}
This work presents VISTA, a monocular global localization framework designed for teams of autonomous vehicles operating within the same environment under extreme appearance changes. VISTA enables vehicles to localize within environment maps generated onboard other vehicles allowing for aligning reference frames to facilitate effective information sharing. This framework features lightweight object-based mapping and efficient correspondence search designed for open-set operation, making it well-suited for collaborative autonomous localization. Experiments demonstrate robust localization performance in the presence of seasonal variation, camera orientation differences, viewpoint variation, lighting changes, occlusions, and perceptual aliasing, highlighting the strengths of segmentation-based tracking and geometric search in unstructured environments.


%%%%%%%%%%%%%%%%%%%%%%%%%%%%%%%%%%%%%%%%%%%%%%%%%%%%%%%%%%%%%%%%%%%%%%%%%%%%%%%%

%%%%%%%%%%%%%%%%%%%%%%%%%%%%%%%%%%%%%%%%%%%%%%%%%%%%%%%%%%%%%%%%%%%%%%%%%%%%%%%%

%%%%%%%%%%%%%%%%%%%%%%%%%%%%%%%%%%%%%%%%%%%%%%%%%%%%%%%%%%%%%%%%%%%%%%%%%%%%%%%%

%%%%%%%%%%%%%%%%%%%%%%%%%%%%%%%%%%%%%%%%%%%%%%%%%%%%%%%%%%%%%%%%%%%%%%%%%%%%%%%%

\bibliographystyle{IEEEtran}
\bibliography{IEEEabrv,bibtex/bib/bibliography}

\end{document}